\documentclass[runningheads]{llncs}
\usepackage[T1]{fontenc}
\usepackage{graphicx}
\usepackage{comment}
\usepackage{booktabs}
\usepackage{pifont}
\usepackage{tikz}
\usetikzlibrary{positioning}
\usepackage{amsmath,amssymb}
\usepackage{tikz}
\usepackage{multirow}
\usetikzlibrary{positioning,arrows.meta}

\makeatletter
\renewcommand\section{\@startsection{section}{1}{\z@}%
	{0.8ex \@plus 0.8ex \@minus 0.8ex}%
	{0.8ex \@plus 0.8ex}%
	{\normalfont\large\bfseries}}
\renewcommand\subsection{\@startsection{subsection}{2}{\z@}%
	{0.7ex \@plus 0.7ex \@minus 0.7ex}%
	{0.7ex \@plus 0.7ex}%
	{\normalfont\normalsize\bfseries}}
\renewcommand\subsubsection{\@startsection{subsubsection}{3}{\z@}%
	{0.7ex \@plus 0.7ex \@minus 0.7ex}%
	{0.7ex \@plus 0.7ex}%
	{\normalfont\normalsize\itshape}}
\makeatother

\begin{document}
\title{From Learning Resources to Competencies: LLM-Based Tagging with Evidence and Graph Constraints}
\titlerunning{LLM-Based Competency Tagging with Evidence and Graph Constraints}
% If the paper title is too long for the running head, you can set
% an abbreviated paper title here
%

%\author*[1,2]{}\email{ngoc-luyen.le@hds.utc.fr}

%\author[2]{\fnm{}\email{marie-helene.abel@hds.utc.fr}
%\equalcont{These authors contributed equally to this work.}

%\author[1,3]{\fnm{Bertrand} \sur{Laforge}}\email{laforge@lpnhe.in2p3.fr}
%\equalcont{These authors contributed equally to this work.}

%\orcidID{0000-1111-2222-3333}
\author{Ng\d{o}c Luy\d{\^{e}}n L\^e\inst{1,2} \and
Marie-H\'el\`ene Abel\inst{2} \and
Bertrand Laforge\inst{3}}
\authorrunning{NL Le et al.}
% First names are abbreviated in the running head.
% If there are more than two authors, 'et al.' is used.
%
\institute{Gamaizer, 93340 Le Raincy, France \and
Universit\'{e} de technologie de Compi\`egne, CNRS, Heudiasyc (Heuristics and Diagnosis of Complex Systems), CS 60319 - 60203 Compi\`egne Cedex, France
 \and
Sorbonne Universit\'{e}, CNRS UMR 7585, LPMHE (Laboratoire de Physique Nucl\'{e}aire et des Hautes \'{E}nergies), 75252 Paris cedex 05, France\\
}
\maketitle              % typeset the header of the contribution
\begin{abstract}
Linking learning resources to a structured competency framework is key to enabling competency-based search and curriculum analytics in Learning Management Systems (LMS). However, manual tagging is labor-intensive, and fully automatic methods often lack transparency. In this paper, we present an end-to-end alignment pipeline that uses a large language model (LLM) as a constrained, evidence-producing tagger. LMS resources -- both instructional content and assessments -- are first segmented into meaningful pedagogical fragments. For each fragment, a small set of candidate competencies is retrieved from structured competency profiles enriched with graph-based context. The LLM then selects the most relevant competencies from this set and provides supporting evidence spans from the fragment text. These predictions are refined using the structure of the competency graph and aggregated at the resource level. We evaluate our approach on a dataset built from the Computer Science department’s competency referential at the \textit{Université de Technologie de Compiègne} (UTC), covering 22 competencies across multiple course materials. Our \textit{LLM+BM25+Graph} (LBG) pipeline achieves strong results, with a micro-F1 of 0.57 and macro-F1 of 0.50 at the fragment level, 0.51 macro-F1 at the resource level, and an MRR of 0.82 -- outperforming zero-shot and few-shot LLM variants, retrieval/similarity baselines, and supervised classifiers -- while also producing more mechanically traceable evidence spans to support human auditing and educational analysis.


\end{abstract}
\section{Introduction}

Competency-based education and curriculum mapping aim to make learning more transparent by explicitly aligning program learning outcomes with course content, learning activities, and assessments. In higher education, curriculum mapping is widely used for program assessment and refinement, but it is often maintained as matrices or maps that require continuous manual updates as courses evolve~\cite{plaza2007curriculum,smith2024defining}.

%Competency-based education and curriculum mapping aim to make learning more transparent and coherent by explicitly aligning program learning outcomes with course content, learning activities, and assessments. This alignment supports constructive alignment in course design, enables instructors to verify coverage and progression of outcomes across a program, and helps institutions document achievement for accreditation, quality assurance, and learner advising. In higher education, curriculum mapping is therefore widely used for program assessment and refinement; however, in practice it is often maintained as static matrices or ad hoc maps that are costly to build and require continuous manual updates as courses evolve, materials are revised, and assessments change~\cite{plaza2007curriculum}. As a result, mappings can quickly become incomplete or outdated, limiting their usefulness for ongoing analytics and decision-making.

In practice, operationalizing the link between an institution’s competency framework and its learning materials remains challenging in Learning Management System (LMS) environments. Course resources are heterogeneous (pages, PDFs, slides, assignments, quizzes) and frequently revised, making outcome-to-content alignment labor-intensive and difficult to keep consistent across instructors and courses. This motivates automated support for curriculum analytics, such as detecting competency coverage gaps and checking coherence between intended outcomes, instructional activities, and assessments~\cite{goff2015learning}.

Recent Large Language Models (LLMs) have shown promise for educational competency/skill and standards tagging, yet performance can vary substantially across referentials and languages~\cite{tang2025llm4tag,le2025well,luyen2025automated}. In particular, models tend to perform better on widely represented frameworks, whereas tagging against local or less represented competency referentials often degrades unless the target framework is explicitly provided or the model is adapted~\cite{kwak2024bridging}. These limitations motivate approaches that (i) constrain the prediction space (e.g., selecting from retrieved competency candidates rather than unconstrained generation) and (ii) return evidence spans to support reliable human review.

From an application perspective, competency alignment is not only a back-office curriculum mapping task: it directly supports everyday LMS workflows. When instructors upload or update materials, automatic suggestions of the relevant competencies can reduce annotation effort and improve consistency at scale, while evidence snippets enable quick validation and correction. Symmetrically, learners can benefit from competency-indexed navigation, for example by finding resources that strengthen a target competency, revising prerequisites before an assessment, or identifying materials that address missing competencies revealed by analytics. These use cases require methods that are accurate, fine-grained, and auditable, rather than opaque predictions.

%In parallel, institutions increasingly require interoperability between local competency frameworks and standard skill referentials used in labor-market and training services. ESCO provides an EU-wide, machine-readable classification of skills/competences and occupations designed to connect education and employment services~\cite{european2019esco}. In France, the ROME 4.0 ecosystem offers open datasets and services that structure occupations and associated competency groups, enabling integration into matching and training pathways. Interoperability therefore raises an additional requirement: mappings should be portable beyond a single institution and exportable to external services.

We address the following problem: given (i) an institution-defined competency graph (including hierarchical and prerequisite relations) and (ii) a set of heterogeneous educational resources, how can we produce accurate resource--competency mappings that are traceable through evidence spans. To this end, we propose an end-to-end pipeline that operates at the fragment level, constrains LLM decisions to plausible candidates, and exploits graph structure for reconciliation before aggregating predictions at the resource level. 

The remainder of this paper is organized as follows. Section~\ref{sec:relatedwork} reviews prior work on competency tagging of educational resources and recent LLM-based approaches, including multilingual settings. Section~\ref{sec:method} presents our approach, an LLM-based competency alignment pipeline with retrieval-constrained candidate selection, evidence spans, and graph-aware reconciliation. Section~\ref{sec:exp} describes the dataset, experimental protocol, and results. Section~\ref{sec:discussion} discusses implications for LMS applications, model and fragmentation choices, and limitations. Section~\ref{sec:conclusion} concludes with directions for future work.

\section{Related Work}\label{sec:relatedwork}
We review work on automatically linking LMS resources to competency frameworks\footnote{We use \emph{competency} to denote a node in our institutional framework, covering what is often termed a \emph{skill}, and reflecting its hierarchical/prerequisite structure.}. We focus on three areas: (i) representing and fragmenting learning resources for alignment, (ii) educational skill tagging methods, and (iii) recent LLM-based competency tagging, including multilingual settings.

%\subsection{Learning resources and competency frameworks}
In LMSs, \emph{learning resources} include instructional and assessment artifacts (e.g., pages, PDFs, quizzes, assignments) that can be analyzed at the resource level and, when needed, segmented into finer units such as sections, slides, or quiz items. A \emph{competency framework} is a structured set of competencies, often organized as a hierarchy/graph and sometimes associated with proficiency levels. Frameworks exist at multiple scopes, from institutional referentials to national and international taxonomies such as \textit{ROME 4.0}~\cite{rome40} and \textit{ESCO}~\cite{european2019esco}. While these layers improve interoperability, they also raise alignment challenges due to differences in granularity, terminology, and language.

%\subsection{Competency tagging in educational content}
Competency tagging has long been studied as \emph{Knowledge Component} (KC) labeling in Intelligent Tutoring Systems, where the main challenge is reducing the cost of manual annotation. Early approaches frame tagging as supervised text classification or nearest-neighbor retrieval over a fixed label space~\cite{karlovvcec2012knowledge}, and later work shows that incorporating contextual signals can improve KC imputation~\cite{pardos2017imputing}. Related work on aligning open educational resources (OER) to standards and taxonomies explores scalable tagging pipelines for resource repositories~\cite{li2024aligning}, although it typically does not leverage course-specific prerequisite/hierarchy graphs.

%\subsection{LLM-based competency tagging}
LLMs have recently been explored for educational skill tagging, but existing results show that performance is sensitive to the target taxonomy and language. Kwak and Pardos quantify cross-country and cross-language disparities in LLM skill tagging and demonstrate that model performance depends on how well the target standards are represented or made explicit to the model~\cite{kwak2024bridging}. These findings motivate careful evaluation and, in practice, constrained decision settings rather than unconstrained label generation.
LLMs have also been applied to assessment-centric tagging. Moore et al.\ use GPT-4 to generate and tag KCs for multiple-choice questions and evaluate agreement with human KC annotations as well as expert preferences in disagreements \cite{moore2024automated}. This work supports the view that assessments are high-signal resources for competency tagging, while also emphasizing that automatic tags must be validated and that top-$k$ recommendation behavior is a realistic evaluation protocol.

%Recent studies show that LLMs can support competency/KC tagging, but performance varies with the target taxonomy and language. Kwak and Pardos report cross-language and cross-standard disparities, motivating constrained settings rather than unconstrained label generation~\cite{kwak2024bridging}. Moore et al.\ apply GPT-4 to generate and tag KCs for multiple-choice questions and emphasize the importance of validation and realistic top-$k$ behavior~\cite{moore2024automated}. Together, these findings motivate evidence-grounded and auditable designs for LMS-scale competency alignment.

In summary, prior work provides strong baselines for mapping educational text to skill labels, while recent LLM studies highlight both potential and limitations. Our work builds on these directions by combining pedagogically motivated fragmentation with retrieval-constrained, evidence-producing LLM tagging and graph-aware reconciliation.

\section{Approach: LLM-Based Competency Tagging Pipeline}\label{sec:method}
In this section, we first formalize the competency \emph{alignment} problem for heterogeneous LMS resources. We then present an end-to-end alignment pipeline in which an LLM acts as a constrained, evidence-producing tagger: for each fragment, it selects competencies from a bounded candidate set and returns verifiable evidence spans. Predictions are subsequently reconciled using the competency graph structure and aggregated at the resource level to support downstream LMS applications.

\subsection{Problem Formulation}
\label{subsec:problem}
Let $G=(C,E)$ denote a competency framework represented as a directed labeled graph, where $C$ is the set of competency nodes and $E$ is the set of relations (e.g., \emph{broader/narrower}, \emph{part-of}, \emph{prerequisite-of}). Let $R$ be a collection of learning resources associated with a course or program, including content-oriented materials (pages, PDFs, slides) and assessment-oriented materials (quizzes, assignments, rubrics). Our objective is to compute a \emph{resource-level} competency mapping
\begin{equation}
	M: R \rightarrow 2^C,
\end{equation}
where $M(r)$ is the set of competencies aligned with resource $r$.
To support human audit, we also compute a \emph{fragment-level} mapping with evidence:
\begin{equation}
	M_f: X \rightarrow 2^{C},\quad X=\bigcup_{r\in R}\phi(r),
\end{equation}
where $\phi(r)$ decomposes each resource into a set of fragments (analysis units) and each predicted label is accompanied by an evidence span from the fragment text.

For example, consider a course \texttt{ML101} with a competency framework graph $G=(C,E)$ where $C=\{c_1:$ $Linear$ $Algebra,c_2:$ $Probability,c_3:$ $Supervised$ $Learning,c_4:$ $Linear$ $Regression,c_5:$ $Logistic$ $Regression$ $/$ $Classification\}$.
Let $E$ include prerequisite relations such as $
E=\{(c_1$ $\rightarrow$ $c_4)$\ $\text{prerequisite-of},$\ $(c_2$ $\rightarrow$ $c_3)$\ $\text{prerequisite-of},$\ $(c_3$ $\rightarrow$ $c_4)$\ $\text{prerequisite-of},$\ $(c_3$ $\rightarrow$ $c_5)$\ $\text{prerequisite-of}\}.
$
Intuitively, \textit{Linear Regression} depends on \textit{Supervised Learning} and mathematical foundations, while \textit{Classification} also depends on \textit{Supervised Learning}.

Let $R$ contain four learning resources from the LMS:  $r_1$ a PDF on ``$Supervised$ $Learning$'', $r_2$ a PDF on ``$Overfitting$'',
$r_3$ a quiz ``$Overfitting$'', and $r_4$ an assignment ``$Classification$''.
Each resource is decomposed into fragments $\phi(r)$, for instance:
\begin{itemize}
	\item $\phi(r_1)=\{x_{1,1}=\text{Section\:1:\:Introduction}, x_{1,2}=\text{Section\:2:\:Problem\:setting}\},$
	\item $\phi(r_2)=\{x_{2,1}=\text{Chapter\:1:\:Overview}, x_{2,2}=\text{Section:\:bias--variance\:intuition}\},$
	\item $\phi(r_3)=\{x_{3,1}=\text{Quiz:\:Overfitting}, x_{3,2}=\text{Quiz:\:mitigation\:strategies}\},$
	\item $\phi(r_4)=\{x_{4,1}=\text{Assignment:\:multi-label\:classification}\}.$
\end{itemize}
Thus $X=\bigcup_{r\in R}\phi(r)$ denotes the set of all fragments.
At the fragment level, the mapping $M_f$ assigns competencies with evidence spans, for example:
\begin{itemize}
	\item $M_f(x_{1,2})$ = $\{(c_3,$\ $\text{``learn\:$ $a\:$ $function\:$ $from\:$ $labeled\:$ $examples''})\},$
	\item  $M_f(x_{2,2})$ = $\{(c_3,$\ $\text{``generalization\:$ $error''}),$\ $(c_5,$\ $\text{``regularization''})\},$
	\item $M_f(x_{4,1})$ = $\{(c_5,$\ $\text{``multi-label\:$ $classification''})\}.$
\end{itemize}

Finally, fragment predictions are aggregated into resource-level mappings, e.g.,
$
M(r_1)=\{c_3\},\quad M(r_2)=\{c_3,c_5\},\quad M(r_3)=\{c_3\},\quad M(r_4)=\{c_5\}.
$
This example illustrates the difference between resource-level alignment $M$ and evidence-grounded
fragment-level alignment $M_f$. It also shows how prerequisite relations in $G$ can be used for coherence
checks (e.g., predicting \textit{Linear Regression} $c_4$ without any supporting evidence for
\textit{Supervised Learning} $c_3$ within the same resource can be flagged as potentially incoherent).

\begin{figure}[t]
	\centering
	\includegraphics[width=\textwidth]{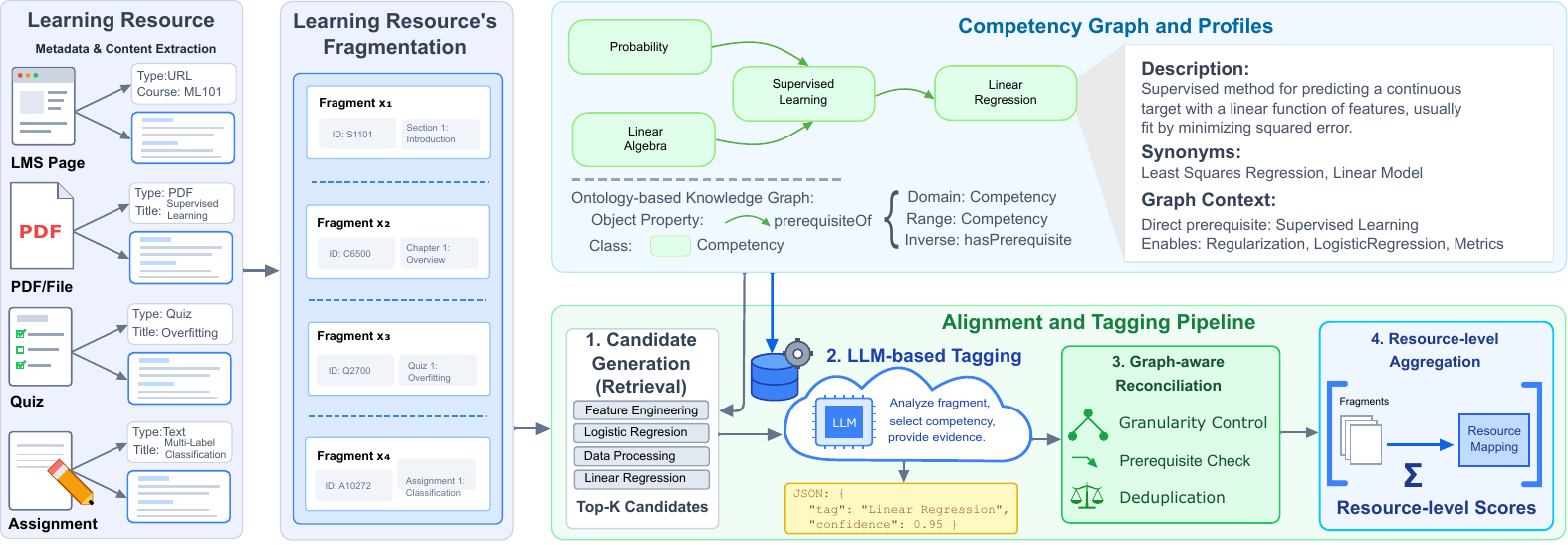}
\caption{Pipeline overview: resource ingestion and fragmentation, candidate retrieval, LLM-based tagging, graph-aware reconciliation, and resource-level aggregation.}

	\label{fig:method-pipeline}
	\vspace{-0.5cm}
\end{figure}

To instantiate the formulation above, we implement the alignment process as an end-to-end pipeline that maps learning resources to resource-level competency scores for evaluation. Figure~\ref{fig:method-pipeline} summarizes the workflow: each resource $r\in R$ is first ingested and represented through metadata and extracted text, then decomposed into fragments $\phi(r)$ to form the set of analysis units $X$. For each fragment $x\in X$, we retrieve a bounded candidate set $Cand(x)\subseteq C$ from competency profiles and apply an LLM to select the relevant competencies within this set while extracting verifiable evidence spans from the fragment text. The resulting fragment-level predictions are subsequently reconciled using structural relations in the competency graph $G$ and aggregated to produce the resource-level mapping $M(r)$. The following subsections detail each stage of the pipeline.

\subsection{Representing Learning Resources}\label{subsec:resource-repr}
Each learning resource $r \in R$ is represented by (i) metadata and (ii) textual content. Metadata includes resource type (e.g., LMS page, file/PDF, URL, quiz, assignment), course identifier, title, and URL. Text is extracted using resource-type-specific routines: the visible HTML body for pages, text extraction for PDFs, and question stems/options/feedback for quizzes and assignments. In LMS environments such as Moodle, this heterogeneous resource structure is explicit and motivates resource-level analysis rather than only course-level descriptions.

Fragmentation into competency-alignment units: Because a single resource often covers multiple topics and competencies, we operationalize each resource as a set of \emph{analysis units} (fragments or chunks) suitable for competency alignment:
\begin{equation}
	\phi(r) = \{x_1,\dots,x_n\}.
\end{equation}
We adopt a fragmentation strategy: chunk boundaries follow instructor-authored pedagogical structure whenever available (e.g., sections in pages/PDFs, slide boundaries,
quiz-item boundaries). This is consistent with prior educational tagging work that treats items (e.g., problems/questions) as natural labeling units \cite{karlovvcec2012knowledge,pardos2017imputing}.
If structure is absent or yields overly long segments, we apply fallback segmentation using paragraph and sentence boundaries with a maximum token budget. Each fragment retains provenance (resource id, section
title, offsets), enabling later auditing of predicted tags and evidence.

\subsection{Competency Graph and Profiles} \label{subsec:competency-profiles}
For each competency $c\in C$, we construct a \emph{competency profile} that serves as the knowledge base for alignment. A profile includes: (i) label(s) and description, (ii) synonyms or alternate labels if available, (iii) structural context in $G$ (e.g., parent/children, prerequisites, related nodes), and (iv) optional examples (typical tasks, keywords). Representing competencies with graph context is aligned with ontology-based curriculum analysis that explicitly models prerequisites and hierarchical relations \cite{nazyrova2023analysis,milosz2024ontological}. Profiles reduce ambiguity and provide a robust basis for retrieval-based candidate selection.

\subsection{Alignment and Tagging Pipeline} \label{subsec:tagging-pipeline}
Given a fragment $x$, we predict a set of competency tags $\hat{T}(x)\subseteq C$ together with confidence and evidence. The pipeline consists of four stages:

\textit{1) Candidate generation (retrieval):} We retrieve the top-$K$ candidate competencies for $x$ by comparing $x$ against competency profiles (e.g., lexical ranking such as BM25, or embedding similarity). Retrieval constrains the label space and improves robustness by avoiding unconstrained label invention. This design is consistent with retrieval-augmented approaches that use retrieval to ground and constrain generation/selection for knowledge-intensive tasks \cite{lewis2020retrieval}.

\textit{2) Competency selection with evidence (LLM core step):}
For each fragment, a tagging model selects zero or more competencies among the candidates and returns: (i) selected competency IDs, (ii) a confidence score, and (iii) an evidence span from $x$ supporting each selected competency. Outputs follow a strict JSON schema to facilitate automatic parsing, evaluation, and auditing. Evidence-centered outputs are supported by work on rationales/evidence as a mechanism to assess model faithfulness and interpretability \cite{deyoung2020eraser}. We use an LLM in this stage because recent studies show LLMs can perform educational skill/KC tagging, while also exhibiting taxonomy- and language-dependent variability that motivates constrained settings and careful evaluation \cite{kwak2024bridging,moore2024automated}.

The LLMs are used as a \emph{constrained semantic decision module}. Conditioned on a fragment $x$ and a retrieved candidate set $Cand(x)$, it performs (i) competency selection among candidates, (ii) confidence estimation, and (iii) evidence extraction (rationale spans) from the fragment. Optionally, the LLM can be used offline to propose alternate labels/synonyms for competency profiles, and as a weak-labeling assistant during annotation; however, our primary use is evidence-grounded selection under a bounded label space.

%\textit{3) Graph-aware reconciliation:}
%The fragment-level predictions produced by the LLM are post-processed using structural constraints from the competency graph $G$. Concretely, we apply reconciliation rules to (i) control granularity by reducing redundant parent--child co-assignments, (ii) assess prerequisite coherence by computing indicators that flag potentially inconsistent tags (rather than forcing automatic corrections), and (iii) encourage parsimony by limiting the number of tags per fragment and per resource. This use of explicit graph structure follows ontology-based curriculum modeling practices, where hierarchies and prerequisites support systematic analysis and consistency checking~\cite{nazyrova2023analysis,milosz2024ontological}.

\textit{3) Graph-aware reconciliation:}
We reconcile the LLM’s fragment-level predictions using structural signals from the competency graph $G$. This reconciliation performs three functions. First, \emph{granularity control} reduces redundant co-assignments between related nodes (e.g., parent--child tags) to favor a consistent level of specificity. Second, we conduct \emph{prerequisite coherence checks} by deriving indicators from prerequisite relations in $G$ to flag potentially incoherent tags (e.g., an advanced competency predicted without supporting evidence for its prerequisites), without forcing automatic corrections. Third, we apply \emph{deduplication} to merge repeated or overlapping predictions of the same competency that may arise across fragments and to remove duplicate tags prior to resource-level aggregation. Such graph-informed post-processing is consistent with ontology-based curriculum modeling, where hierarchies and prerequisites are exploited for analysis and consistency checking~\cite{nazyrova2023analysis,milosz2024ontological}.

%\textit{4) Resource-level aggregation:} Fragment-level tags are aggregated into resource-level tags:
%\begin{equation}
%	score(r,c) = \mathrm{Agg}\left(\{conf(x,c): x\in\phi(r)\}\right),
%\end{equation}
%where $\mathrm{Agg}$ can be $\max$ or a weighted sum depending on fragment type (e.g., assessment fragments such as quizzes and rubric criteria receive higher weights). We output the top-$k$ %competencies per resource subject to a threshold $\tau$. The resulting mappings support downstream curriculum analytics, such as competency coverage and gap detection \cite{gottipati2018competency}.

\textit{4) Resource-level aggregation:}
We convert fragment-level predictions into a resource-level competency scoring function. For each resource $r\in R$, let $\phi(r)=\{x_1,\dots,x_{n_r}\}$ denote its fragments. The tagging stage yields, for each fragment $x$ and competency $c\in C$, a confidence score $conf(x,c)\in[0,1]$ (with $conf(x,c)=0$ if $c\notin \hat{T}(x)$ after reconciliation). We define the resource-level score of competency $c$ for resource $r$ as a weighted aggregation over fragments:
\begin{equation}
	s(r,c)=\mathrm{Agg}\big(\{w(x)\,conf(x,c)\ :\ x\in\phi(r)\}\big),
	\label{eq:resource-score}
\end{equation}
where $w(x)$ is an optional fragment-type weight and $\mathrm{Agg}$ is either $\max$ (evidence of $c$ in any fragment is sufficient) or a weighted sum/mean (accumulated evidence across fragments). The resource-level mapping is then obtained by thresholding and optional $\mathrm{TopK}$ selection:
\begin{equation}
	M(r)=\{c\in C \;:\; s(r,c)\ge \tau\},\qquad
	M_k(r)=\mathrm{TopK}\big(\{(c,s(r,c))\}_{c\in C},\,k\big).
	\label{eq:resource-mapping}
\end{equation}
%The resulting resource--competency matrix (Figure~\ref{fig:method-pipeline}) supports downstream curriculum analytics such as competency coverage estimation and gap detection~\cite{gottipati2018competency}.

Overall, our approach supports competency-based analysis by organizing heterogeneous resources into a set of representations: (i) the competency graph $G$ enriched with competency profiles, (ii) a learning resource with extracted text and metadata, (iii) a fragment corpus $\phi(r)$ with provenance, and (iv) evidence-grounded alignments linking fragments and resources to competencies. 
About the \emph{Downstream Artifacts/Applications}, these representations enable both systematic evaluation -- including tagging accuracy, evidence validity/faithfulness, and graph-coherence indicators -- and practical LMS applications such as competency-based retrieval and navigation, assessment alignment and quality control, and curriculum coverage analysis~\cite{gottipati2018competency}. The next section details our experimental setup, datasets, baselines, and evaluation metrics.

%\section{Experimentation}\label{sec:exp}

%Overall, our approach yields a set of reusable artifacts: (i) a competency graph dataset (nodes/edges) augmented with competency profiles, (ii) a learning-resource corpus with extracted text and metadata, (iii) a fragment corpus $\phi(r)$ with provenance information, and (iv) evidence-grounded mappings between fragments/resources and competencies at both fragment and resource levels. These outputs enable systematic evaluation of tagging quality (accuracy, faithfulness, and graph coherence) and support downstream uses such as competency-based retrieval and curriculum coverage analytics \cite{gottipati2018competency}. In the next section, we describe our experimental setup, datasets, baselines, and evaluation metrics used to assess the proposed approach. 

%We evaluate the proposed evidence-grounded alignment pipeline (Figure~\ref{fig:method-pipeline}) on a real-world learning-resource corpus aligned with the Informatics competency framework of the Université de Technologie de Compiègne (UTC). 
\section{Experimentation}\label{sec:exp}
Our experimental study addresses three research questions: \textit{RQ1}~to what extent does the method accurately align LMS fragments and resources to competencies; \textit{RQ2}~what is the contribution of retrieval-based candidate constraints and graph-aware reconciliation to alignment quality; and \textit{RQ3}~how mechanically verifiable are the evidence spans produced by the model.

\subsection{Dataset}\label{subsec:dataset} 
%We adopt the UTC Informatics competency referential, modeled as a directed competency graph $G=(C,E)$ with $|C|=22$ competencies spanning multiple levels of granularity, including parent--child relations. Each competency is represented by a profile containing a French label and description and, when available, English labels or aliases to support bilingual prompting. As summarized in Table~\ref{tab:dataset-stats}, the corpus covers 26 course units (UVs) and includes 430 learning resources. Structure-first fragmentation yields $N$ fragments in total; we annotate a subset of 432 fragments with gold competency identifiers for evaluation.

We adopt the competency referential of the Computer Science Department at the \textit{Université de Technologie de Compiègne} (UTC), modeled as a directed competency graph $G=(C,E)$ with $|C|=22$ competency nodes spanning multiple levels of granularity (including parent--child relations). Each node is associated with a competency profile comprising a French label and description and, when available, English labels or aliases used to support bilingual prompting. Table~\ref{tab:dataset-stats} reports corpus statistics: the dataset covers 26 course units (UVs) and includes 430 resources. For evaluation, 432 fragments are annotated with gold competency identifiers.

To construct the corpus, we work at the UV level. For each UV, we begin with the official course description and associated LMS entry page, which provides a canonical list of learning materials and activities, along with links to any external platforms hosting supplementary content. We collect these items, normalize their metadata (UV, type, title, source), and extract their text content using resource-specific parsing routines. Finally, we segment the content into pedagogically meaningful fragments.

For evaluation, gold annotations are created at the fragment level. Resource-level gold label sets are then derived deterministically from these annotations: for each resource $r$, the gold resource label set is defined as the union of the gold competency labels assigned to its annotated fragments, i.e.,
\[
Y^{\mathrm{res}}_r = \bigcup_{x \in \phi(r)} Y_x.
\]
No additional graph-based reconciliation or thresholding is applied to the gold labels. In other words, graph-aware reconciliation is used only on the prediction side, whereas the resource-level gold set is an aggregation of manual fragment-level annotations.

\begin{table}[t]
	\centering
	\begin{tabular}{lr}
		\hline
		Statistic & Value \\
		\hline
		Number of competencies ($|C|$) & 22 \\
		Number of UVs (course units) & 26 \\
		Number of resources & 430 \\
		%Number of fragments (all) & 432 \\
		Number of gold fragments (annotated) & 432 \\
		Median fragment length (chars) & 624 \\
		\hline
	\end{tabular}
	\caption{Dataset statistics for the UTC's Computer Science Department competency alignment task.}
	\label{tab:dataset-stats}
	\vspace{-1cm}
\end{table}

\subsection{Experimental Setting}\label{subsec:protocol}

We evaluate cross-course generalization with 5-fold UV-level cross-validation: all fragments from the same UV are kept in the same fold to avoid content overlap between train and test. This setting matches deployment, where alignment must transfer to unseen courses.

For retrieval-constrained variants, we tune two parameters: the number of candidates $K\in\{5,10,15,20\}$ retrieved per fragment and a confidence threshold $\tau\in\{0.3,0.4,0.5,0.6\}$. The retrieval parameter $K$ bounds the label space presented to the LLM, while $\tau$ controls which predicted labels are retained during filtering and/or aggregation. Hyperparameter selection is performed within the training portion of each fold (inner selection), and evaluation is conducted on the corresponding held-out fold.

%The constrained labeling stage uses \texttt{gpt-4o-mini} and follows an evidence-producing protocol~\cite{lewis2020retrieval,deyoung2020eraser}. Given a fragment and its candidate set, the model selects one or more competency identifiers exclusively from the candidates, assigns a confidence score, and highlights a supporting text span from the fragment. Model outputs are required to follow a machine-readable structure to enable automatic parsing and validation. We then verify that reported character offsets refer to a valid substring of the fragment; malformed outputs are discarded.

The constrained labeling stage uses \texttt{gpt-4o-mini} and follows an evidence-producing protocol~\cite{lewis2020retrieval,deyoung2020eraser}. Given a fragment and its candidate set, the model selects one or more competency identifiers exclusively from the candidates, assigns a confidence score, and highlights a supporting text span from the fragment. Model outputs are required to follow a machine-readable structure to enable automatic parsing and validation. We then verify that reported character offsets refer to a valid substring of the fragment; malformed outputs are discarded. These confidence scores are used as operational ranking/filtering signals rather than interpreted as calibrated probabilities.

%The returned confidence scores are used operationally as ranking and filtering signals during aggregation rather than interpreted as calibrated probabilities. Accordingly, our sensitivity analysis studies the effect of the confidence threshold $\tau$ on downstream tagging performance, while a full calibration analysis is left for future work.

%The constrained labeling stage is instantiated with \texttt{gpt-4o-mini} and follows an evidence-producing rationale protocol~\cite{lewis2020retrieval,deyoung2020eraser}. Given a fragment and its candidate set, the model is instructed to (i) select one or more competency identifiers exclusively from the candidate set, (ii) assign a confidence score to each selected competency, and (iii) provide a supporting evidence span as a verbatim quote with character offsets. Outputs are required to follow a strict JSON schema. We enforce validity by parsing the JSON and verifying that each offset pair $(a,b)$ references a defined substring of the fragment text; malformed outputs are discarded.

%\subsection{Compared Methods}\label{subsec:methods}

\subsection{Compared Methods}\label{subsec:methods}
We compare our designed pipeline, \textit{LLM+BM25+Graph} (LBG), against retrieval, retrieval \& similarity, LLM-based, and supervised baselines:

%\textbf{Our pipeline - }: retrieves a bounded candidate set with BM25 over competency profiles, applies a constrained LLM to select competencies and extract evidence spans, then performs graph-aware reconciliation before resource-level aggregation.

\textbf{LLM baselines:} \textit{Zero-shot LLM} which prompts the LLM to tag directly against the full competency inventory; and (ii) \textit{Few-shot LLM}, which adds in-context demonstrations from training UVs but still omits graph reconciliation~\cite{brown2020language}.

 \textbf{Retrieval and similarity baselines}: We evaluate candidate-generation methods: \textit{BM25}~\cite{robertson2009probabilistic}, \textit{Dense} (bi-encoder)~\cite{karpukhin2020dense}, \textit{Hybrid--RRF}~\cite{cormack2009reciprocal}, and \textit{BM25+Cross-Encoder}~\cite{nogueira2019passage}. \textit{SBERT similarity}~\cite{reimers2019sentence} selects the top-$K$ competencies by cosine similarity. Ranked lists are converted to multi-label predictions by selecting top-$K$ per fragment (fixed $K$ across folds) prior to aggregation. We note that this fixed top-$K$ conversion may disadvantage ranking-based baselines relative to LBG, which additionally uses a confidence threshold $\tau$ on the LLM outputs. We therefore interpret these retrieval-only results primarily as candidate-ranking baselines rather than fully optimized multi-label decision systems.

\textbf{Supervised baselines: } We train \textit{LogReg+TF--IDF}~\cite{wang2012baselines}, \textit{LinearSVC+TF--IDF}~\cite{joachims1998text} and a \textit{Supervised Transformer}~\cite{sanh2019distilbert} for multi-label competency prediction.

\subsection{Evaluation Metrics}\label{subsec:metrics}

We evaluate competency alignment as a multi-label prediction task over the competency set $C$. For each unit (fragment or resource) $u$, let $Y_u\subseteq C$ be the gold labels and $\hat{Y}_u\subseteq C$ the predicted labels.

\textbf{Micro-F1 (fragment level):}
Let $U$ denote the set of evaluated units. The pooled counts are
$
\mathrm{TP}=\bigl|\{(u,c)\in U\times C : c\in Y_u \ \wedge\ c\in \hat{Y}_u\}\bigr|,\quad
\mathrm{FP}=\bigl|\{(u,c)\in U\times C : c\notin Y_u \ \wedge\ c\in \hat{Y}_u\}\bigr|,\quad
\mathrm{FN}=\bigl|\{(u,c)\in U\times C : c\in Y_u \ \wedge\ c\notin \hat{Y}_u\}\bigr|.
$

\[
F1_{\mu}=\frac{2\,\mathrm{TP}}{2\,\mathrm{TP}+\mathrm{FP}+\mathrm{FN}}.
\]

\textbf{Macro-F1 (fragment level):}
For each competency $c\in C$, let
$
\mathrm{TP}_c = \bigl|\{u : c\in Y_u \ \wedge\ c\in \hat{Y}_u\}\bigr|,\quad
\mathrm{FP}_c = \bigl|\{u : c\notin Y_u \ \wedge\ c\in \hat{Y}_u\}\bigr|,\quad
\mathrm{FN}_c = \bigl|\{u : c\in Y_u \ \wedge\ c\notin \hat{Y}_u\}\bigr|.
$
\[
F1_c=\frac{2\,\mathrm{TP}_c}{2\,\mathrm{TP}_c+\mathrm{FP}_c+\mathrm{FN}_c},\qquad
F1_{\mathrm{macro}}=\frac{1}{|C|}\sum_{c\in C}F1_c.
\]
Micro-F1 weights frequent competencies more heavily, whereas macro-F1 weights all competencies equally.

\textbf{Resource Macro-F1:}

We compute the same $F1_{\mathrm{macro}}$ after aggregating fragment predictions into a single label set per resource (i.e., $u$ ranges over resources rather than fragments), yielding $F1^{\mathrm{res}}_{\mathrm{macro}}$. Resource-level gold labels are constructed as the union of manually annotated fragment-level gold labels for the corresponding resource.

\textbf{Evidence span validity:}
For span-producing methods, we measure mechanical validity of the returned offsets. For each fragment $i$ with text $x_i$, the model outputs spans $\hat{S}_i$ with character offsets $(a,b)$. A span is valid if it forms a non-empty interval within the fragment:
\[
\mathrm{valid}(a,b)=
\begin{cases}
	1 & \text{if } 0 \le a < b \le |x_i|,\\
	0 & \text{otherwise.}
\end{cases}
\qquad
\mathrm{SpanValid}=\frac{\sum_i\sum_{(a,b)\in \hat{S}_i}\mathrm{valid}(a,b)}{\sum_i|\hat{S}_i|}.
\]
This metric captures whether evidence can be automatically traced back to the original text; it does not evaluate whether the evidence semantically justifies the predicted label.

\textbf{Mean Reciprocal Rank (MRR):}
We measure candidate-ranking quality with MRR. For each fragment $i$, a method outputs a ranked list of competencies and a gold label set $G_i$. We define the rank of the first correct competency as
$
	r_i=\min\{k\ge 1 \mid \mathrm{ranked}_i[k]\in G_i\},
$
and set $RR_i=1/r_i$ if a gold competency appears in the top-$K$, otherwise $RR_i=0$. The mean reciprocal rank is
$
	MRR=\frac{1}{N}\sum_{i=1}^{N} RR_i,
$
where $N$ is the number of evaluated fragments. Higher MRR indicates that a correct competency is ranked earlier.

%Table~\ref{tab:main-results} reports the main results averaged over 5-fold UV-level cross-validation. Across all metrics, \textbf{LBG} provides the strongest performance, outperforming both LLM baselines and the retrieval-only baselines. Few-shot prompting combined with BM25 candidates improves over the zero-shot LLM setting, indicating that in-context demonstrations help calibrate competency selection. However, the full pipeline remains superior, consistent with the complementary role of retrieval constraints and graph-based reconciliation. Keyword matching and BM25-only remain substantially below the LLM-based approaches, suggesting that lexical retrieval alone is insufficient to reliably map pedagogical fragments to competency identifiers.
	\begin{comment}
	\begin{tabular}{lcccc}
		\hline
		Method & micro-F1 & macro-F1 & res.\ macro-F1 & Evid.\ valid \\
		\hline
		LBG (best $K,\tau$) & 0.57 & 0.50 & 0.51 & 0.51 \\
		Few-shot LLM    & 0.49 & 0.40 & 0.41 & 0.41 \\
		Zero-shot LLM                  & 0.45 & 0.37 & 0.37 & 0.39 \\
		BM25-only                      & 0.23 & 0.22 & 0.22 & -- \\
		Keyword baseline               & 0.35 & 0.30 & 0.30 & -- \\
		%	SentenceBERT               & 0.146 & 0.143 & 0.075 & -- \\
		%	SVM + TF-IDF                   & 0.078 & 0.052 & 0.053 & -- \\
		\hline
	\end{tabular}
	
\end{comment}

\subsection{Results}\label{subsec:results}

Table~\ref{tab:main-results} reports UV-level cross-validation results for all compared methods. Regarding \textit{RQ1 (alignment accuracy)}, \textit{LBG} achieves the best performance across fragment-level Micro-F1 and Macro-F1, as well as the best resource-level macro-F1 after aggregation. This indicates that the improvements are not limited to local fragment decisions, but remain beneficial when predictions are summarized at the resource level, which is the granularity typically used in LMS applications.

%For \textit{RQ2 (contribution of candidate constraints and graph-aware reconciliation)}, we observe a consistent progression from \textit{Zero-shot LLM} to \textit{Few-shot LLM} to \textit{LBG}. Few-shot prompting improves over zero-shot, suggesting that in-context demonstrations help the model interpret pedagogical phrasing and map it to competency identifiers. The additional gain of \textit{LBG} supports the role of retrieval-based candidate bounding and graph-aware post-processing as complementary mechanisms: restricting predictions to BM25-retrieved candidates reduces spurious labels and promotes specificity, while reconciliation regularizes outputs by controlling granularity, performing deduplication, and flagging prerequisite incoherence using relations in $G$.

For \textit{RQ2 (contribution of candidate constraints and graph-aware reconciliation)}, the current results provide preliminary evidence that progressively constraining the decision process improves alignment quality: performance increases from \textit{Zero-shot LLM} to \textit{Few-shot LLM} and then to \textit{LBG}. However, this progression alone does not fully isolate the contribution of retrieval-based candidate restriction and graph-aware reconciliation as separate components. To address this point more directly, targeted ablations remain part of our ongoing work; in the present paper, the reported comparisons provide preliminary evidence for the complementary role of retrieval constraints and graph-aware reconciliation.

%For \textit{RQ2 (contribution of candidate constraints and graph-aware reconciliation)}, the current results provide preliminary evidence that progressively constraining the decision process improves alignment quality: performance increases from \textit{Zero-shot LLM} to \textit{Few-shot LLM} and then to \textit{LBG}. However, this progression alone does not fully isolate the contribution of retrieval-based candidate restriction and graph-aware reconciliation as separate components. To address this point more directly, we include targeted ablations comparing the retrieval-constrained LLM with and without graph-aware reconciliation, and discuss the distinct role of each reconciliation function.

The expanded set of \textit{retrieval \& similarity baselines} clarifies the limits of ranking-only approaches under our evaluation protocol. BM25, Dense, and hybrid retrievers obtain low F1 when converted to multi-label predictions via top-$K$ selection, indicating that retrieving plausible candidates alone is insufficient without an explicit semantic decision rule for multi-label assignment. In contrast, \textit{BM25+Cross-Encoder} yields substantially stronger F1 than other retrieval-only methods, demonstrating that supervised reranking can mitigate some ambiguity in candidate selection. Nevertheless, it remains below \textit{LBG} on both fragment- and resource-level F1, highlighting the added value of constrained LLM selection coupled with graph-based reconciliation. \textit{SBERT similarity} exhibits relatively high MRR but limited resource-level F1, suggesting that while relevant competencies may appear early in the ranked list, top-$K$ conversion still introduces many false positives.

\begin{table}[t]
	\centering
	
	\begin{tabular}{llccccc}
	\toprule
	Group & Method & Micro-F1 & Macro-F1 & $F1^{\mathrm{res}}_{\mathrm{macro}}$ & SpanValid & MRR \\
	\midrule	
	\multirow{5}{6.0em}{Retrieval \& similarity-based}& BM25 & 0.0770 & 0.0758 & 0.0762 & -- & 0.3407 \\
	&Dense (Bi-encoder) & 0.0806 & 0.0799 & 0.0804 & -- & 0.7357 \\
	&Hybrid RRF & 0.0773 & 0.0767 & 0.0770 & -- & 0.5641 \\
	%&Hybrid Weighted & 0.0773 & 0.0767 & 0.0771 & -- & 0.5963 \\
	&BM25+Cross Encoder & 0.3990 & 0.2766 & 0.2794 & -- & 0.7552 \\
	&SBERT Similarity & 0.1459 & 0.1434 & 0.0745 & -- & 0.7357 \\
	\hline
	\multirow{3}{6.0em}{Supervised-based}& LogReg + TF-IDF & 0.1189 & 0.1174 & 0.1187 & -- & 0.3917 \\
	&LinearSVC + TF-IDF & 0.1218 & 0.1201 & 0.1211 & -- & 0.3462 \\
	&Supervised Transformer & 0.1158 & 0.1125 & 0.1136 & -- & 0.2306 \\
	%&SVM + TF-IDF & 0.0649 & 0.0400 & 0.0418 & -- & 0.2981 \\
	\hline

	\multirow{3}{6.0em}{LLM-based}& Zero-shot LLM & 0.4482 & 0.3689 & 0.3698 & 0.3877 & 0.6020 \\
	&Few-shot LLM & 0.4858 & 0.4047 & 0.4071 & 0.4120 & 0.7031 \\
	
	&LBG (best $K$,$\tau$) & 0.5710 & 0.5037 & 0.5072 & 0.5107 & 0.8247 \\
	\bottomrule
	\end{tabular}

	\caption{Overall results, evidence validity is reported only for span-producing LLM methods; other methods are not applicable (``--'').}
	\label{tab:main-results}
	\vspace{-1cm}
\end{table}

We operate at fragment level because many LMS resources mix several competencies within a single document, while the final LMS-facing output remains resource-level through aggregation.

The \textit{supervised baselines} (TF--IDF linear models and a supervised Transformer) underperform the LLM-based systems in this cross-UV setting. This suggests that, given the size and diversity of the annotated fragment corpus, purely supervised classifiers generalize less effectively across course units than approaches that leverage retrieval grounding and evidence-driven selection.

We report \textit{MRR} to evaluate ranking quality. Dense retrieval and cross-encoder reranking achieve relatively high MRR, indicating that a correct competency is often ranked early in the candidate list. However, high MRR does not necessarily translate into high F1 when rankings are converted into multi-label predictions via top-$K$ selection: a method can rank a gold competency early while still including many non-gold competencies among the top positions, increasing false positives. \textit{LBG} achieves the highest MRR overall, consistent with its design: BM25 provides strong candidates, while the constrained LLM stage filters irrelevant items and promotes earlier placement of truly supported competencies.

Finally, addressing \textit{RQ3 (evidence validity)}, \textit{LBG} yields the highest span validity among span-producing methods. This suggests that bounding the label space and stabilizing predictions through graph-aware reconciliation also stabilizes evidence extraction, improving the mechanical traceability of rationales and supporting human audit and downstream LMS use.

\begin{figure}[t]
	\centering
	\includegraphics[width=0.88\linewidth]{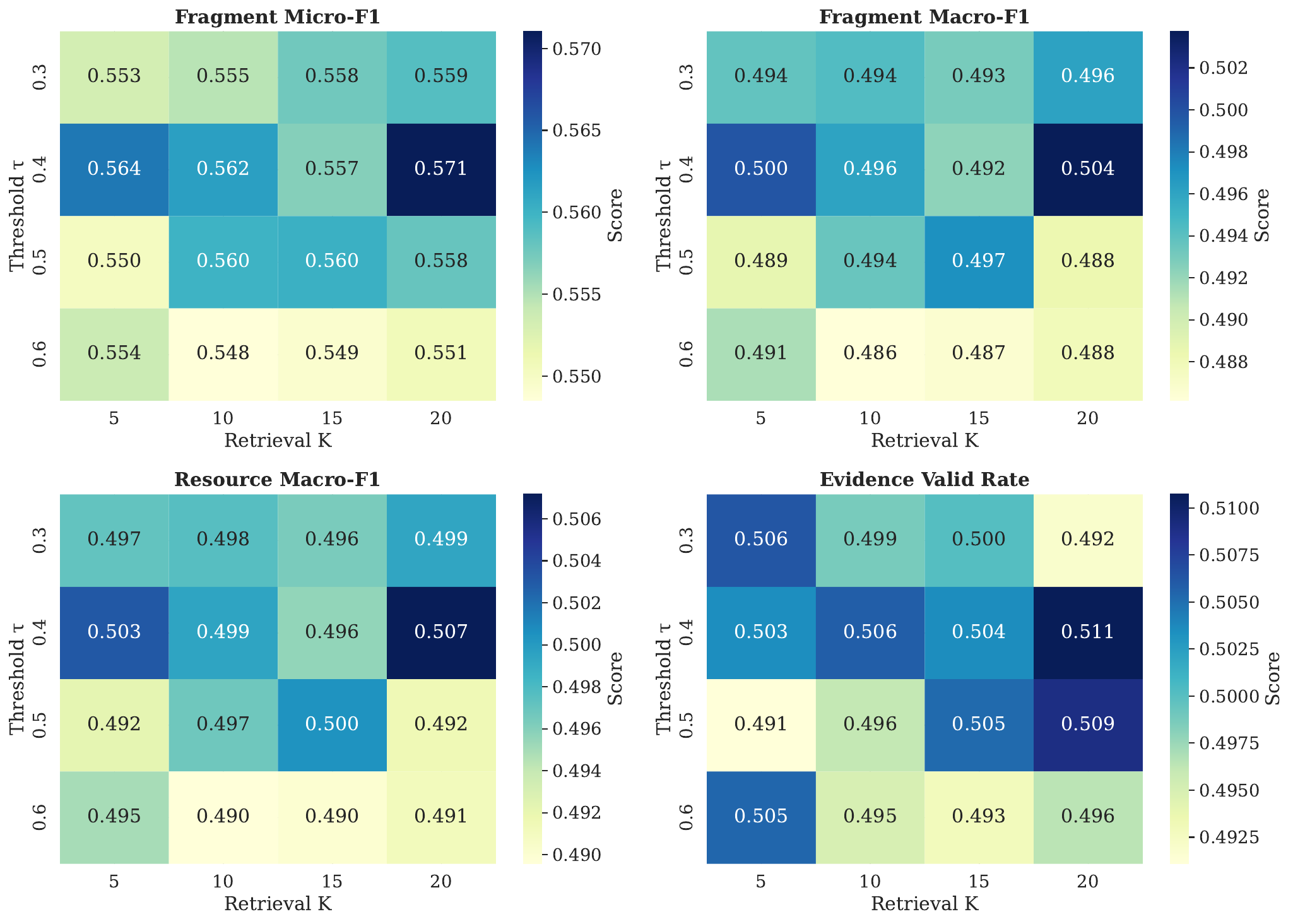}
	\caption{Sensitivity of \textbf{LBG} to retrieval depth $K$ and confidence threshold $\tau$.}
	\label{fig:metrics-heatmap}
\vspace{-0.5cm}
\end{figure}

\textit{Sensitivity to retrieval depth $K$ and threshold $\tau$ (Figure~\ref{fig:metrics-heatmap}):}
The heatmap summarizes the effect of the retrieval depth $K$ and confidence threshold $\tau$ on fragment-level performance and evidence validity. Overall, performance is relatively stable across the explored grid: micro-F1 varies within $[0.548,\,0.571]$ and macro-F1 within $[0.486,\,0.504]$, indicating that the approach is not overly sensitive to small hyperparameter changes. The best operating point remains $K{=}20$ and $\tau{=}0.4$, achieving micro-F1 $=0.571$, macro-F1 $=0.504$, and resource macro-F1 $=0.507$, together with the highest evidence validity ($0.511$). Increasing $K$ generally improves results by exposing the LLM to a richer candidate set, whereas $\tau$ controls a precision--recall trade-off: for $K{=}20$, moving from $\tau{=}0.3$ to $\tau{=}0.4$ increases both precision and recall (micro-precision $0.469\rightarrow0.479$, micro-recall $0.740\rightarrow0.758$), while larger thresholds reduce performance as correct low-confidence labels are increasingly filtered.
% Sensitivity heatmap (K, tau) — use your generated file name
\section{Discussion}\label{sec:discussion}
Our results demonstrate that combining retrieval-constrained LLM tagging with graph-based reconciliation yields consistent gains over unconstrained LLM baselines and traditional retrieval or classification methods. The strongest performance across F1 and MRR metrics is achieved by the LBG pipeline, confirming the value of bounding the label space and enforcing structural consistency via the competency graph. In particular, improvements in evidence span traceability suggest that constraining the prediction space also helps stabilize the LLM’s rationale generation, which is critical for transparency and traceability for human audit in educational settings.

%An important design choice in our framework is the fragmentation of resources into pedagogically meaningful units. This enables fine-grained tagging and localized evidence, but also makes performance sensitive to fragment granularity. Coarse fragments may mix multiple competencies and introduce label noise, whereas overly fine fragments may lack sufficient context for reliable tagging. Our current rule-based segmentation uses existing LMS structure; exploring adaptive or learning-based fragmentation strategies is a promising direction for improving precision without sacrificing coverage.

An important design choice in our framework is the fragmentation of resources into pedagogically meaningful units. We operate at fragment level because many LMS resources mix several competencies within a single document, while the final LMS-facing output remains resource-level through aggregation. This enables fine-grained tagging and localized evidence, but also makes performance sensitive to fragment granularity. Coarse fragments may mix multiple competencies and introduce label noise, whereas overly fine fragments may lack sufficient context for reliable tagging. Our current rule-based segmentation uses existing LMS structure; exploring adaptive or learning-based fragmentation strategies is a promising direction for improving precision without sacrificing coverage.

Another factor concerns the choice of LLM backend. We employ \texttt{gpt-4o-mini}, a lightweight model selected to balance computational cost and latency; nevertheless, when coupled with retrieval-based candidate restriction and graph-aware reconciliation, it consistently outperforms the non-LLM baselines. It is reasonable to expect that larger or domain-adapted models could further improve F1 and evidence quality, especially for subtle or implicitly expressed competencies. This suggests that our pipeline design is model-agnostic: performance can scale with stronger LLMs while preserving the interpretability benefits of evidence-grounded, graph-constrained tagging.

Beyond offline evaluation, our outputs can be used directly in an LMS: resource-competency scores enable competency-based retrieval and navigation with evidence highlights, support assessment alignment by flagging mismatches between taught and assessed competencies, and feed coverage dashboards that summarize coverage and gaps at course or program level. This helps instructors audit and organize materials, and helps learners find relevant resources and track progress through evidence-backed competency views.

Finally, our evaluation is limited to a single institutional framework (UTC's Computer Science Department) and text-only resources. Generalizing to larger and multilingual taxonomies such as \textit{ESCO} or \textit{ROME} will likely require stronger candidate generation and richer graph constraints to handle greater label ambiguity and scale. Nevertheless, our outputs -- competency-indexed resources, evidence-linked fragments, and resource-competency matrices -- already support practical LMS applications such as competency-based retrieval and navigation, assessment alignment, and curriculum coverage analytics.

\section{Conclusion}\label{sec:conclusion}
In this paper, we presented an end-to-end pipeline for aligning heterogeneous LMS resources with an institution-defined competency graph by combining fragmentation, retrieval-constrained LLM tagging with evidence spans, and graph-aware reconciliation prior to resource-level aggregation. On our experimental dataset, \textit{LBG} achieves the best fragment- and resource-level F1 and the highest MRR among all compared methods, outperforming zero-/few-shot LLM variants, retrieval/similarity baselines, and supervised classifiers; it also yields the most mechanically valid evidence spans, supporting traceability for human audit. Limitations include evaluation on a single institutional framework, an evidence metric that tests offset traceability rather than semantic faithfulness, and reliance on competency profile text for retrieval. Moving forward, we will focus on (i) validating across additional frameworks and institutions, (ii) strengthening evidence assessment with human faithfulness checks, and (iii) improving candidate generation and reconciliation with hybrid/dense representations and more principled graph constraints for downstream LMS applications such as competency-based retrieval and assessment alignment.

\begin{credits}
\subsubsection{\ackname}
We warmly thank the Ikigai consortium led by the association Games for Citizens, the company Gamaizer, as well as the FORTEIM project (winner of the AMI CMA France 2030 call for projects), for their support and collaboration. Their contributions have provided significant added value to the completion of this research.

\subsubsection{Data \& code availability.} 
The dataset and implementation code for the proposed pipeline are publicly available at \url{https://github.com/lengocluyen/competency-tagging}.
\end{credits}
%
% ---- Bibliography ----
%
% BibTeX users should specify bibliography style 'splncs04'.
% References will then be sorted and formatted in the correct style.
%
\bibliographystyle{splncs04}
\bibliography{references}
\end{document}